\pdfoutput=1 
\documentclass[11pt]{article}
\usepackage{EACL2023}
\usepackage{times}
\usepackage{latexsym}
\usepackage[T1]{fontenc}
\usepackage[utf8]{inputenc}
\usepackage{microtype}
\usepackage{inconsolata}
\usepackage{booktabs}
\usepackage{multirow}
\usepackage[capitalise,nameinlink]{cleveref}
\usepackage{enumitem}

\newcommand{\da}{da}
\newcommand{\de}{de}
\newcommand{\en}{en}
\newcommand{\deit}{de-IT}
\newcommand{\dech}{de-CH}
\newcommand{\itit}{it}
\newcommand{\itna}{it-NA}
\newcommand{\nl}{nl}
\newcommand{\ro}{ro}
\newcommand{\romd}{ro-MD}

\newcommand{\es}{es}
\newcommand{\escr}{es-CR}
\newcommand{\esmx}{es-MX}
\newcommand{\espe}{es-PE}
\newcommand{\esuy}{es-UY}

\let\origpm\pm
\renewcommand{\pm}{\,\origpm}

\title{Fine-Tuning BERT with Character-Level Noise for \\ Zero-Shot Transfer to Dialects and Closely-Related Languages}

\author{Aarohi Srivastava \and David Chiang \\
  University of Notre Dame, USA \\ 
  \texttt{\{asrivas2,dchiang\}@nd.edu}}

\begin{document}
\maketitle

\begin{abstract}
In this work, we induce character-level noise in various forms when fine-tuning BERT to en-\\able zero-shot cross-lingual transfer to unseen dialects and languages. We fine-tune BERT on three sentence-level classification tasks and evaluate our approach on an assortment of unseen dialects and languages. We find that character-level noise can be an extremely effective agent of cross-lingual transfer under certain conditions, while it is not as helpful in others. Specifically, we explore these differences in terms of the nature of the task and the relationships between source and target languages, finding that introduction of character-level noise during fine-tuning is particularly helpful when a task draws on surface level cues and the source-target cross-lingual pair has a relatively high lexical overlap with shorter (i.e., less meaningful) unseen tokens on average. 
\end{abstract}

\section{Introduction} \label{sec:intro}

Contemporary NLP methods such as BERT \cite{devlin2019bert}, with the large amount of knowledge contained within their parameters, paired with the relatively low computational power required to fine-tune them for a downstream task, have taken over many NLP applications.
Indeed, several monolingual and multilingual BERT models are available that encompass a number of languages \cite{devlin2019bert}.
However, the strength of these models is tied to the availability of data, and the large amounts of data required to pre-train such models exclude some languages for which it is difficult to collect large amounts of written text. 

The scarcity of data becomes more severe with dialects and language varieties.  In fact, the very nature of dialects as an evolving form of the language, often spoken rather than written, with various social and cultural nuances, can make it difficult to develop systems tailored to specific dialects. 
In many applications, users may span a continuum of idiolects, some falling into established dialects and others not.
It may therefore be impossible to train a system on even a small amount of data in every idiolect.
In this paper, we consider \emph{zero-shot cross-lingual transfer}, which we define strictly as any scenario in which the test data is of a language variety not included in any stage of training.  For instance, we may fine-tune a standard Italian BERT model on standard Italian sentiment analysis data, and then perform inference on Neapolitan (a variety closely related to standard Italian). We call standard Italian the \emph{source} language, and Neapolitan the \emph{target} language.

The mismatch in BERT's performance when evaluating on the source versus target language can arise for a variety of reasons depending on the properties of the target language.  For example, some language varieties may have similar morphology but different vocabulary, so that BERT may encounter completely new words when tested on the dialect. An example of this is the use of ``soda'' in some regions of the United States and ``pop'' in others to refer to a carbonated beverage; a model trained only on the ``soda'' varieties may have difficulty identifying the meaning of ``pop'' if it appears in test data. 

In other cases, language varieties may have similar vocabulary, but phonological, morphological, or orthographic differences may throw off the subword tokenization method used by the model. A simple example is the distinction in spelling between ``color'' (American English) and ``colour'' (British English); if a model were trained exclusively on American English and ``colour'' was not part of its vocabulary, at test time, ``colour'' would be tokenized differently than ``color,'' possibly resulting in a different interpretation by the model.

In this work, we focus on the second type of dialectal variation. Following \citet{aepli-sennrich-2022-improving}, we study how introducing character-level noise in training can improve performance for zero-shot cross-lingual transfer between closely-related languages.
\citeauthor{aepli-sennrich-2022-improving}'s \citeyearpar{aepli-sennrich-2022-improving} method is a two-step process. The first step is \emph{continued pre-training} of BERT on three types of data: target language data, un-noised source language data, and noised source language data. The second step is \emph{fine-tuning} on noised task data in the source language. Here, we only use fine-tuning, and we only use source-language data, making our method strictly zero-shot.
We explore the next questions: which techniques of character-level noising help cross-lingual transfer the most, and in which situations should one expect character-level noising to work best?

To explore these questions, we fine-tune monolingual BERT models on three sentence-level classification tasks: intent classification, topic identification, and sentiment analysis. We introduce multiple variations on the method of noising in order to optimize cross-lingual transfer.  We test our methods on an assortment of unseen languages, some closely-related and some more distant relatives.  For the intent classification task, our systems work almost perfectly, that is, they perform nearly as well on the target language as on the source language. We also boost task performance in less closely-related languages (in the same and different families). Furthermore, we find that we can obtain even bigger improvements by using more noise, and we find that exposing the model to more variations of the data during fine-tuning also helps.  Finally, we explore the conditions for cross-lingual transfer needed for our method to be successful.

\section{Background and Related Work}

\subsection{Fine-tuning BERT for Dialectal NLP}
There are many previous findings that fine-tuning a BERT model on a specific task involving dialectal data leads to high performance on the task with dialectal test data. Examples include sentiment analysis on Arabic dialects \cite{abdel-salam-2022-dialect, fsih-etal-2022-benchmarking, husain-etal-2022-weak}, hate speech detection for Egyptian-Arabic \cite{10009167}, part-of-speech tagging for North-African Arabizi \cite{srivastava:hal-02270527}, and sentiment analysis for Hong Kong Chinese \cite{10020704}. The success in diverse applications of the general method informs our decision to stay within the paradigm of BERT fine-tuning; however, without task-labeled fine-tuning data available in the test dialect/language, we must do something else in the fine-tuning step (in our case, inducing character-level noise) in order to facilitate zero-shot cross-lingual transfer.

\subsection{Adversarial Learning}
Adversarial learning has been employed in the space of zero-shot cross-lingual transfer with success \cite{ponti2018adversarial, huang2019cross, he2020adversarial, dong2020leveraging}.  However, this line of work draws on additional learning techniques and/or model architectures (e.g., BiLSTMs and GANs), expending extra computation for training, rather than working within the scope of fine-tuning; additionally, adversarial attacks are often done in the embedding space rather than to the words themselves.  At the same time, it provides an intuition that inclusion of adversarial examples in training can be an effective tool in various applications.

\subsection{Zero-Shot Cross-Lingual Transfer}

In \cref{sec:intro}, we gave a narrow definition of zero-shot transfer as using no target-language data at all during training.
For example, fine-tuning standard Italian BERT on standard Italian, then testing on Neapolitan, would meet our definition.
Note that it is possible for the pre-training data of Italian BERT to contain Neapolitan text given that the pre-training data is constructed by scraping various online sources \cite{devlin2019bert}; however, because the presence of Neapolitan text would likely be accidental in the pre-training, we do not control for this. In contrast, if the source language is Italian and the target language is Spanish, and we were to use multilingual BERT as the pre-trained model, we would not consider this zero-shot cross-lingual transfer, as multilingual BERT includes Spanish as one of the intentional training languages. 

All past work in zero-shot cross-lingual transfer that we are aware of has used some target-language data during training, whether by using a multilingual model or by introducing new data from the target language at some stage. 
Approaches involving meta-learning \cite{nooralahzadeh2020zero} and adapter layers \cite{vidoni2020orthogonal,parovic2022bad} add a component to the model and train it specifically to the target language.  Under the BERT-based paradigm, \citet{wang2019cross} learn contextual word alignments to align the contextualized embeddings in the source and target language, which requires parallel text in the source and target languages. \Citet{tian2021rumour} fine-tune BERT in the source language, then generate ``silver labels'' in the target language and iteratively fine-tune on those; although this doesn't require parallel data, it still requires target-language data. \Citet{huang-etal-2021-improving-zero} employ adversarial training and randomized smoothing for zero-shot cross-lingual transfer; though their method does not introduce additional data from the target language during training, they work with multilingual models that include the target languages in the pre-training. As described above (\cref{sec:intro}), the method \citet{aepli-sennrich-2022-improving} use is directly related to ours, but is not strictly zero-shot, because continued pre-training uses target-language data.

\section{Methods}
Our experiments focus on fine-tuning monolingual language models on same-language data, and testing for zero-shot cross-lingual transfer to other languages, inducing noise in the fine-tuning data to facilitate this transfer.  Building on \citeauthor{aepli-sennrich-2022-improving}'s \citeyearpar{aepli-sennrich-2022-improving} promising finding that character-level noise can be used as a conduit for cross-lingual transfer between closely-related languages, we introduce a range of options for applying character noise in order to explore how we can better leverage the benefits of character-level noise for zero-shot cross-lingual transfer. 

\subsection{Model}
The models we use in our experiments are all BERT-type models \citep{devlin2019bert} with one additional fine-tuning layer for sentence-level classification.   We use the base size (12 Transformer encoder layers) of the relevant monolingual BERT models for our tasks, topped with a linear classifier which maps the start-of-sentence \texttt{CLS} token to a sentence-level class.  In our setup, the pre-trained model is fine-tuned on one of three sentence-level classification tasks: intent classification, topic identification, and sentiment analysis.  All models used are the uncased versions for simplicity and consistency. In an effort to minimize computation and stick to the zero-shot case, a distinction we make from \citeauthor{aepli-sennrich-2022-improving}'s \citeyearpar{aepli-sennrich-2022-improving} work is to limit experiments to fine-tuning only (no continued pre-training). 

\subsection{Noising Technique}
Our noising technique is similar to that of \citet{aepli-sennrich-2022-improving}. We begin with raw text, and we define a word to be any continuous substring of letters (identified using Python's \texttt{isalpha} function). For each word, with probability $p$ we apply noise to the word, and with probability $1-p$ we leave the word unchanged. We leave non-words (for example, numbers, symbols, and punctuation) unchanged, as we expect variation between closely related languages to primarily affect words.  We express $p$ as a percentage and refer to it as the \emph{noise level}.  Noise is applied at a single, randomly selected character position in the word, meaning that noise can only be applied to a word up to one time.  

We include four possible types of character-level noise in the fine-tuning data. Three are in common with \citet{aepli-sennrich-2022-improving}: \emph{insertion}, \emph{deletion}, and \emph{replacement}.  We also add \emph{swapping} between adjacent letters. We describe the noising technique below (Section~\ref{noisingvariations}). All four of these operations are present in cases of dialectal variation. For example, American English spells words like \textit{color} with an \textit{or} ending, while the British English spelling has an insertion of \textit{u} as in \textit{colour} (or vice versa, there is a deletion from British to American English).  Metathesis results in swapping adjacent sounds (sometimes realized in orthography), such as \textit{ask} in standard English and \textit{aks} in some varieties.

As insertion and replacement require inclusion of an additional character outside those in the word, the character is chosen from the alphabet of the language of the noised text.  For example, if the text to apply noise to is in English, the alphabet would consist of letters \emph{a} through \emph{z}, while for German, the alphabet would also consist of umlaut vowels (\emph{ä}, \emph{ö}, \emph{ü}) and the eszett (\emph{ß}).  All random selections are uniform within the set of possibilities. Below, we exemplify how each type of noise may be applied by taking the example of the word \emph{straw}:
\setlist{nolistsep}
\begin{itemize}[noitemsep]
    \item \emph{Insert} a randomly selected alphabet letter (``j'') at a randomly selected index of the word (index 1): \emph{sjtraw}.
    \item \emph{Delete} the letter at a randomly selected index of the word (index 2): \emph{staw}.
    \item \emph{Replace} the letter at a randomly selected index of the word (index 3) with a randomly selected alphabet letter (``o''): \emph{strow}.
    \item \emph{Swap} the letter at a randomly selected index of the word (not including the final index of the word) with the subsequent letter of the word: \emph{strwa}.
\end{itemize}

\subsection{Noising Variations}
\label{noisingvariations}
\citet{aepli-sennrich-2022-improving} used 10--15\% character-level noise in their fine-tuning data and found their method to be effective in promoting cross-lingual transfer.  Given the promise of their result, we introduce two dimensions along which to vary the noise application: noise level and composition of fine-tuning data.  In addition to the baseline (0\% noise level), we employ higher levels of noise: 25\%, 50\%, 75\%, and 100\% of words.

Because the goal is to expose BERT to different spellings and tokenizations of the same word during fine-tuning, we include multiple copies of the fine-tuning data, each with some difference in noise.  The more copies we include, the more we might expect the model to adapt to surface-level variation in the context of the task.

We tried two possible compositions: \emph{joint} and \emph{stacked}.  In the \emph{joint} composition, we include two copies of the fine-tuning data: the first copy is the original data without noise, and the second copy is noised using all four types of noise in equal proportion.  In the \emph{stacked} composition, we include five copies of the fine-tuning data: the first copy is, once again, the original data without noise, and the remaining copies are noised with each of the four types of noise, respectively.  Including multiple copies allows the model to see the same sentences during fine-tuning with variations in spelling (and thereby the token sequence). 

For reference, assuming a noise level of 50\%, the compositions would appear as follows:
\setlist{nolistsep}
\begin{itemize}[noitemsep]
    \item Joint-composition:
    \setlist{nolistsep}
    \begin{enumerate}[noitemsep]
        \item Original data (0\% noise level)
        \item Noised data: 12.5\% each of insertion, deletion, replacement, and swapping noise.
    \end{enumerate}
    \item Stacked-composition:
    \setlist{nolistsep}
    \begin{enumerate}[noitemsep]
        \item Original data (0\% noise level)
        \item Insertion-noised data (50\% noise level)
        \item Deletion-noised data (50\% noise level)
        \item Replacement-noised data (50\% noise level)
        \item Swapping-noised data (50\% noise level)
    \end{enumerate}
\end{itemize}

\section{Experiments}
\label{experiments}
In order to evaluate the effectiveness of inducing character-level noise for zero-shot cross-lingual transfer under the various settings described in Section~\ref{noisingvariations}, we test on three tasks: intent classification, topic identification, and sentiment analysis. All three tasks are sentence-level classification tasks; however, each task has unique challenges that can bolster or break compatibility with our approach.  We are interested in seeing how noise can help in each of these scenarios.

\subsection{Tasks}
The intent classification task we use is xSID \cite{van-der-goot-etal-2021-masked}, a benchmark for cross-lingual slot and intent detection that includes parallel labeled data in 13 languages. The xSID dataset was drawn from the English Snips \cite{coucke2018snips} and cross-lingual Facebook \cite{schuster2019cross} datasets and translated to the other languages.  We take German (\de{}) and Italian (\itit{}) to be the source languages in our experiments; the training data consists of 10,000 sentences each, and the validation data consists of 300 sentences each.  We do not use any data from the target languages until inference; the test data for each language consists of 300 sentences.  There are 18 total intent labels for classification.  For the most part, each sample is a simple imperative or interrogative (e.g., ``Remind me to wake up around 6 am tomorrow.''). Our intent classification system is included in the 2023 VarDial Evaluation Campaign \cite{2023-findings-vardial}. 

The topic identification task we use is MOROCO \cite{butnaru2019moroco}, a Moldavian (\romd{}) and Romanian (\ro{}) dialectal corpus which consists of news text from these two language varieties labeled by topic.  There are five possible topic labels: culture, finance, politics, science, sports, and tech.  There are 21,719 training samples and close to 6,000 validation and test samples each.  In contrast to the intent classification data, MOROCO samples contain much longer multi-sentence text.  In addition, \citet{butnaru2019moroco} remove named entities from the data in order to minimize the ability to use surface level cues to solve the task.  We take Romanian to be the source language and Moldavian to be the target language for our experiments.

The sentiment analysis task we use is TASS 2020 \cite{garcia2020overview}, a Spanish dialectal corpus which consists of tweets from five varieties of Spanish: Spain (\es{}), Costa Rica (\escr{}), Mexico (\esmx{}), Peru (\espe{}), and Uruguay (\esuy{}).  Given that much of the pre-training data for Spanish BERT \citep{jose_canete_2019_3247731, CaneteCFP2020} comes from European sources, we take the Spain subset to be the source language, and the remaining four varieties to be the target languages.  There are three possible sentiment analysis labels: positive, neutral, and negative.  The Spain training subset contains 1126 examples.  For each variety, the test data contains close to 1000 examples.  

\subsection{Fine-Tuning}
For each task, we fine-tune the relevant BERT model on task data from a single source language, and test on other related target languages. We fine-tune each model five times with a different random initialization each time, and report the average across the five trials.  For intent classification, we take German and Italian to be the source languages, fine-tuning German BERT\footnote{\url{https://huggingface.co/dbmdz/bert-base-german-uncased}} on the German subset of xSID and Italian BERT\footnote{\url{https://huggingface.co/dbmdz/bert-base-italian-uncased}} on the Italian subset of xSID.  For topic identification, we take Romanian to be the source of transfer and fine-tune Romanian BERT\footnote{\url{https://huggingface.co/dumitrescustefan/bert-base-romanian-uncased-v1}} \cite{dumitrescu-etal-2020-birth} on the Romanian subset of MOROCO.  For sentiment analysis, we take Spain Spanish to be the source of transfer and fine-tune Spanish BERT \citep{CaneteCFP2020} on the corresponding subset of TASS 2020. 

We fine-tune the baseline model, as well as eight variations to facilitate zero-shot cross-lingual transfer for each task.  Recall that the baseline model is fine-tuned only on data from the source language, and the possible variations are in terms of noise level and composition of fine-tuning data.  The eight variations all involve fine-tuning with noise -- we test all combinations of noise level (25\%, 50\%, 75\%, or 100\%) and composition of fine-tuning data (joint vs.~stacked).  Because the stacked composition includes more copies of the fine-tuning data, we adjust the number of epochs so that each variation is trained for the same number of steps.  Thus, the intent classification and sentiment analysis joint-composition models are fine-tuned for 5 epochs, while the stacked composition models are fine-tuned for 2 epochs.  However, in the topic identification task, we find that training for 2 epochs in the joint-classification model yields better validation performance than 5 epochs, so we train for 2 epochs in both settings of topic identification.  

\subsection{Testing}
We evaluate each model on test data from multiple target languages in order to determine each model's effectiveness in supporting zero-shot cross-lingual transfer.  We also test on the source language to ensure that performance is maintained despite the introduction of noise.  Note that tests are restricted to languages that share the same script as the fine-tuning data. 

For the German intent classification models, we test on 2 dialects of German: Swiss German (\dech{}) and South Tyrolean (\deit{}); 3 Germanic languages (phylogenically closest to farthest): Dutch (\nl{}), English (\en{}), and Danish (\da{}), and 1 non-Germanic language: Italian (\itit{}). For the Italian intent classification models, we test on one dialect of Italian (Neapolitan, \itna{}) and one non-Romance language (German, \de{}). For the Romanian topic identification models, we test on Moldavian. For the Spanish sentiment analysis models, we test on the four Latin American varieties of Spanish included in the TASS 2020 dataset: Costa Rica, Mexico, Peru, and Uruguay. 

The results for our experiments (Section~\ref{experiments}) are presented in Table~\ref{intent-de} (German intent classification), Table~\ref{intent-it} (Italian intent classification), Table~\ref{topic-ro} (topic identification), and Table~\ref{sa-es} (sentiment analysis). Each reported score is the average of five trials and accompanied by the 95\% confidence interval.  Our results demonstrate that our character-level noise intervention boosts performance anywhere from 11 to 40 percentage points across all language pairs tested for intent classification (except English), while maintaining or even raising performance on the source language. We suspect that the approach did not work well for English due to the fact that, unlike the other target languages, English has much more of a loan-word culture, commonly using words from several languages of origin. Curiously, our results also show that the character-level noise intervention was not helpful for the topic identification and sentiment analysis tasks.  Below, we investigate the reasons behind the performance boosts in intent classification as they relate to our noise settings (noise level and composition of fine-tuning data), as well as the differences in the tasks (nature of the task and cross-lingual transfer pairs) that result in such a sharp contrast in the utility of character-level noise.
\section{Results}

\begin{table*}\centering
\small
\begin{tabular}{@{}rlrrrrrrrr@{}}
\toprule
\multicolumn{1}{c}{\textbf{\begin{tabular}[c]{@{}c@{}}Noise\\ Level\end{tabular}}} & \multicolumn{1}{c}{\textbf{\begin{tabular}[c]{@{}c@{}}Comp-\\ osition\end{tabular}}} & \multicolumn{1}{c}{\textbf{\de{}}} & \multicolumn{1}{c}{\textbf{\dech{}}} & \multicolumn{1}{c}{\textbf{\deit{}}} & \multicolumn{1}{c}{\textbf{\nl{}}} & \multicolumn{1}{c}{\textbf{\en{}}} & \multicolumn{1}{c}{\textbf{\da{}}} & \multicolumn{1}{c}{\textbf{\itit{}}} & \multicolumn{1}{c}{\textbf{Average}} \\ \midrule
\textbf{0\%}                                                                       & N/A                                                                                  & 98.2$\pm$0.6                        & 74.9$\pm$7.4                         & 59.5$\pm$8.8                           & 37.0$\pm$3.9                        & 78.0$\pm$1.4                        & 38.8$\pm$4.8                        & 21.3$\pm$1.3                        & 58.2$\pm$1.7                             \\
\textbf{25\%}                                                                      & Joint                                                                                & 97.9$\pm$0.4                        & 71.2$\pm$7.7                         & 67.3$\pm$8.0                           & 37.1$\pm$3.9                        & 74.1$\pm$2.6                        & 39.2$\pm$4.4                        & 25.3$\pm$4.1                        & 58.9$\pm$1.6                             \\
\textbf{50\%}                                                                      & Joint                                                                                & 98.2$\pm$0.8                        & 89.3$\pm$2.6                         & 85.7$\pm$3.3                           & 67.6$\pm$2.0                        & 77.3$\pm$1.9                        & 62.5$\pm$4.2                        & 34.9$\pm$5.6                        & 73.6$\pm$1.6                             \\
\textbf{75\%}                                                                      & Joint                                                                                & 98.7$\pm$0.3                        & 92.7$\pm$1.3                         & 89.9$\pm$2.0                           & 68.5$\pm$2.2                        & \textbf{79.3}$\pm$1.0                        & 61.9$\pm$4.2                        & 34.5$\pm$4.9                        & 75.1$\pm$1.7                             \\
\textbf{100\%}                                                                     & Joint                                                                                & 98.4$\pm$0.5                        & 94.6$\pm$2.5                         & 90.4$\pm$3.9                           & 73.1$\pm$1.3                        & 78.2$\pm$1.3                        & \textbf{65.5}$\pm$2.2                        & \textbf{44.5}$\pm$5.0                        & 77.8$\pm$1.0                             \\
\textbf{25\%}                                                                      & Stacked                                                                              & 98.8$\pm$0.4                        & 91.4$\pm$4.5                         & 86.3$\pm$1.8                           & 58.1$\pm$5.6                        & 77.9$\pm$1.4                        & 56.0$\pm$5.6                        & 28.9$\pm$3.5                        & 71.0$\pm$2.5                             \\
\textbf{50\%}                                                                      & Stacked                                                                              & \textbf{99.0}$\pm$0.4                        & 93.6$\pm$2.7                         & \textbf{91.7}$\pm$2.2                           & 66.4$\pm$1.4                        & 78.0$\pm$3.2                        & 60.4$\pm$2.9                        & 37.1$\pm$3.5                        & 75.2$\pm$1.2                             \\
\textbf{75\%}                                                                      & Stacked                                                                              & 98.7$\pm$0.2                        & 94.1$\pm$2.2                         & 90.3$\pm$2.6                           & 71.2$\pm$5.1                        & 78.0$\pm$1.7                        & 64.2$\pm$3.8                        & 41.9$\pm$4.7                        & 76.9$\pm$2.0                             \\
\textbf{100\%}                                                                     & Stacked                                                                              & \textbf{99.0}$\pm$0.5                        & \textbf{95.3}$\pm$1.7                         & 90.5$\pm$2.9                           & \textbf{77.0}$\pm$2.8                        & 77.5$\pm$1.3                        & 63.5$\pm$2.3                        & 44.4$\pm$2.7                        & \textbf{78.2}$\pm$1.1                             \\ \bottomrule
\end{tabular}
\caption{\label{intent-de}\centering Intent classification results for German BERT with 95\% confidence interval measured for five trials. Bold numbers indicate the highest results (by absolute comparison).}
\end{table*}

\begin{table*}
    \centering
    \small
\begin{tabular}{@{}rlrrrrl@{}}
\toprule
\multicolumn{1}{c}{\textbf{\begin{tabular}[c]{@{}c@{}}Noise\\ Level\end{tabular}}} & \multicolumn{1}{c}{\textbf{\begin{tabular}[c]{@{}c@{}}Comp-\\ osition\end{tabular}}} & \multicolumn{1}{c}{\textbf{\itit}} & \multicolumn{1}{c}{\textbf{\itna}} & \multicolumn{1}{c}{\textbf{\de}} & \multicolumn{2}{c}{\textbf{Average}} \\ \midrule
\textbf{0\%}                                                                       & N/A                                                                                & 97.5$\pm$0.6                        & 79.9$\pm$0.7                         & 31.7$\pm$5.5                        & \multicolumn{2}{r}{69.7$\pm$1.7}         \\
\textbf{25\%}                                                                      & Joint                                                                                & 98.1$\pm$0.3                        & 79.9$\pm$0.4                         & 33.7$\pm$6.0                        & \multicolumn{2}{r}{70.6$\pm$1.8}         \\
\textbf{50\%}                                                                      & Joint                                                                                & 98.0$\pm$0.7                        & 90.3$\pm$0.2                         & 37.0$\pm$2.2                        & \multicolumn{2}{r}{75.1$\pm$0.8}         \\
\textbf{75\%}                                                                      & Joint                                                                                & 97.7$\pm$0.3                        & 91.3$\pm$1.4                         & 42.3$\pm$6.4                        & \multicolumn{2}{r}{77.1$\pm$2.1}         \\
\textbf{100\%}                                                                     & Joint                                                                                & 97.9$\pm$0.5                        & 93.1$\pm$0.5                         & \textbf{45.2}$\pm$3.1                        & \multicolumn{2}{r}{\textbf{78.8}$\pm$1.1}         \\
\textbf{25\%}                                                                      & Stacked                                                                              & \textbf{98.3}$\pm$0.3                        & 90.0$\pm$1.0                         & 34.3$\pm$4.1                        & \multicolumn{2}{r}{74.2$\pm$1.6}         \\
\textbf{50\%}                                                                      & Stacked                                                                              & 97.6$\pm$0.8                        & 93.2$\pm$1.2                         & 43.7$\pm$1.8                        & \multicolumn{2}{r}{78.2$\pm$1.0}         \\
\textbf{75\%}                                                                      & Stacked                                                                              & 97.7$\pm$0.5                        & \textbf{93.4}$\pm$0.5                         & 42.3$\pm$3.3                        & \multicolumn{2}{r}{77.8$\pm$1.2}         \\
\textbf{100\%}                                                                     & Stacked                                                                              & 96.6$\pm$0.8                        & 91.0$\pm$1.1                         & 44.7$\pm$2.5                        & \multicolumn{2}{r}{77.4$\pm$0.8}         \\ \bottomrule
\end{tabular}
\caption{\label{intent-it}\centering Intent classification results for Italian BERT with 95\% confidence interval measured for five trials. Bold numbers indicate the highest results (by absolute comparison).}
\end{table*}

\begin{table}\centering
\small
\begin{tabular}{@{}rlrrr@{}}
\toprule
\multicolumn{1}{c}{\textbf{\begin{tabular}[c]{@{}c@{}}Noise\\ Level\end{tabular}}} & \multicolumn{1}{c}{\textbf{\begin{tabular}[c]{@{}c@{}}Comp-\\ osition\end{tabular}}} & \multicolumn{1}{c}{\textbf{\ro{}}} & \multicolumn{1}{c}{\textbf{\romd{}}} & \multicolumn{1}{c}{\textbf{Average}} \\ \midrule
\textbf{0\%}                                                                       & N/A                                                                                & 77.7$\pm$0.6                        & \textbf{85.7}$\pm$0.8                        & \textbf{81.7}$\pm$0.5                             \\
\textbf{25\%}                                                                      & Joint                                                                                & 77.8$\pm$0.7                        & 82.2$\pm$4.5                        & 80.0$\pm$2.5                             \\
\textbf{50\%}                                                                      & Joint                                                                                & 77.2$\pm$0.7                        & 84.9$\pm$1.9                        & 81.1$\pm$1.3                             \\
\textbf{75\%}                                                                      & Joint                                                                                & 77.7$\pm$0.7                        & 83.1$\pm$2.5                        & 80.4$\pm$1.1                             \\
\textbf{100\%}                                                                     & Joint                                                                                & \textbf{77.9}$\pm$0.8                        & 81.6$\pm$3.8                        & 79.7$\pm$2.1                             \\
\textbf{25\%}                                                                      & Stacked                                                                              & 75.1$\pm$0.5                        & 80.0$\pm$3.5                        & 77.5$\pm$1.9                             \\
\textbf{50\%}                                                                      & Stacked                                                                              & 76.3$\pm$0.5                        & 83.5$\pm$1.6                        & 79.9$\pm$0.9                             \\
\textbf{75\%}                                                                      & Stacked                                                                              & 77.0$\pm$0.5                        & 83.8$\pm$1.5                        & 80.4$\pm$0.9                             \\
\textbf{100\%}                                                                     & Stacked                                                                              & 77.3$\pm$0.5                        & 82.6$\pm$3.7                        & 79.9$\pm$1.7                             \\ \bottomrule
\end{tabular}
\caption{\label{topic-ro}Topic identification results for Romanian BERT with 95\% confidence interval measured for five trials.}
\end{table}

\begin{table*}
    \centering
    \small
\begin{tabular}{@{}rlrrrrrr@{}}
\toprule
\multicolumn{1}{c}{\textbf{\begin{tabular}[c]{@{}c@{}}Noise\\ Level\end{tabular}}} & \multicolumn{1}{c}{\textbf{\begin{tabular}[c]{@{}c@{}}Comp-\\ osition\end{tabular}}} & \multicolumn{1}{c}{\textbf{\es{}}} & \multicolumn{1}{c}{\textbf{\escr{}}} & \multicolumn{1}{c}{\textbf{\esmx{}}} & \multicolumn{1}{c}{\textbf{\espe{}}} & \multicolumn{1}{c}{\textbf{\esuy{}}} & \multicolumn{1}{c}{\textbf{Average}} \\ \midrule
\textbf{0\%}                                                                       & N/A                                                                                  & 66.9$\pm$2.2                        & 62.6$\pm$1.9                        & 66.6$\pm$2.3                        & 49.6$\pm$3.0                        & \textbf{64.4}$\pm$1.9                        & \textbf{61.5}$\pm$1.6                             \\
\textbf{25\%}                                                                      & Joint                                                                                & \textbf{67.3}$\pm$0.7                        & 62.7$\pm$1.3                        & 66.9$\pm$0.6                        & 47.6$\pm$2.0                        & 63.0$\pm$1.0                        & 61.2$\pm$0.5                             \\
\textbf{50\%}                                                                      & Joint                                                                                & 65.8$\pm$1.5                        & \textbf{63.4}$\pm$1.3                        & 66.9$\pm$1.0                        & 49.4$\pm$2.6                        & 64.1$\pm$1.9                        & 61.4$\pm$0.7                             \\
\textbf{75\%}                                                                      & Joint                                                                                & 67.0$\pm$1.7                        & 61.7$\pm$1.8                        & 66.1$\pm$2.7                        & 48.7$\pm$3.0                        & 63.6$\pm$2.0                        & 61.0$\pm$0.7                             \\
\textbf{100\%}                                                                     & Joint                                                                                & 66.3$\pm$1.8                        & 62.5$\pm$1.8                        & 66.0$\pm$1.7                        & \textbf{49.9}$\pm$3.0                        & 63.3$\pm$0.8                        & 61.3$\pm$0.6                             \\
\textbf{25\%}                                                                      & Stacked                                                                              & 66.3$\pm$1.2                        & 61.8$\pm$1.4                        & 66.9$\pm$1.8                        & 49.1$\pm$2.0                        & 63.7$\pm$1.7                        & 61.2$\pm$0.6                             \\
\textbf{50\%}                                                                      & Stacked                                                                              & 66.7$\pm$1.0                        & 62.7$\pm$2.2                        & 66.9$\pm$0.9                        & 47.5$\pm$1.8                        & 63.7$\pm$1.9                        & 61.1$\pm$1.2                             \\
\textbf{75\%}                                                                      & Stacked                                                                              & 66.0$\pm$1.9                        & 63.2$\pm$1.1                        & 66.0$\pm$1.6                        & 48.9$\pm$1.8                        & 64.2$\pm$1.5                        & 61.2$\pm$0.6                             \\
\textbf{100\%}                                                                     & Stacked                                                                              & 67.2$\pm$1.1                        & 61.3$\pm$2.3                        & \textbf{68.4}$\pm$0.7                        & 45.3$\pm$3.0                        & 62.7$\pm$0.9                        & 60.7$\pm$1.1                             \\ \bottomrule
\end{tabular}
    \caption{\centering Sentiment analysis results for Spanish BERT  with 95\% confidence interval measured for five trials. Bold numbers indicate the highest results (by absolute comparison).}
    \label{sa-es}
\end{table*}

\subsection{Level of Noise}
Our intent classification results demonstrate that noise can be an extremely effective tool in promoting zero-shot cross-lingual transfer.  While \citet{aepli-sennrich-2022-improving} use noise levels of 10\% and 15\% in their experiments, we use higher noise levels (25\%, 50\%, 75\%, and 100\%).  Across our intent classification experiments, we find a trend towards ``the more, the better'' when it comes to character-level noise -- 100\% noise is the best (comparing the average scores across all languages).  In German intent classification, for transfer to closely-related varieties (\dech{}) and \deit{}), our intervention is capable of boosting performance very close to the German performance itself.  We also nearly double performance for the other, less closely-related languages tested, including the non-Germanic language tested, Italian, though there is still a big distance to the German performance.  Similarly, in Italian intent classification, we are able to bring Neapolitan performance close to the Italian source performance, and we even raise accuracy for German, a non-Romance language.  

These takeaways from the intent classification task show that character-level noise can be an extremely effective agent for cross-lingual transfer for both close (dialects) and more distant (different families) language pairs.  However, while it is nearly enough to bring about comparable performance for closely related languages, it is of course not enough to do the same for more distant language pairs.

\subsection{Composition of Fine-tuning Data}
In conjunction with using higher levels of noise, we also experiment with two methods of composing the fine-tuning data, which integrate the noise differently (joint vs.~stacked composition).  Recall that under both compositions, one copy of the original fine-tuning data is included; in the joint composition, we include one additional copy that contains four types of character-level noise, while the stacked composition includes \emph{four} additional copies, each with a distinct type of noise.  Having more copies of the data, each with varied spelling and tokenization, allows the model to build robustness to such variation. In addition, though the noise level within each copy of the data is the same, including more copies with noise increases the proportion of noise in the data as a whole (over all the copies).  We find that the stacked-composition models perform better on average than the joint-composition models for the intent classification task (+4 points in German, +1 point in Italian).

\subsection{Lexical Overlap}
\label{sec:lexicaloverlap}
\begin{table}\centering\small
\begin{tabular}{@{}llrr@{}}
\toprule
\textbf{Source} & \multicolumn{1}{c}{\textbf{Target}} & \multicolumn{1}{c}{\textbf{\begin{tabular}[c]{@{}c@{}}Lexical\\ Overlap (\%)\end{tabular}}} & \multicolumn{1}{c}{\textbf{\begin{tabular}[c]{@{}c@{}}Average Length\\ of OOV Tokens\end{tabular}}} \\ \midrule
\textbf{\de{}}     & \de{}                                  & 92.5                                                                                        & 5.3                                                                                                 \\
\textbf{}       & \dech{}                               & 84.7                                                                                        & 4.4                                                                                                 \\
\textbf{}       & \deit{}                               & 89.1                                                                                        & 4.8                                                                                                 \\
\textbf{}       & \nl{}                                  & 83.7                                                                                        & 4.2                                                                                                 \\
\textbf{}       & \en{}                                  & 79.4                                                                                        & 4.3                                                                                                 \\
\textbf{}       & \da{}                                  & 83.7                                                                                        & 4.2                                                                                                 \\
\textbf{}       & \itit{}                                  & 84.1                                                                                        & 4.3                                                                                                 \\
\textbf{\itit{}}     & \itit{}                                  & 93.2                                                                                        & 5.7                                                                                                 \\
\textbf{}       & \itna{}                            & 90.2                                                                                        & 5.2                                                                                                 \\
                & de                                  & 87.7                                                                                        & 4.4                                                                                                 \\
\textbf{\ro{}}     & \ro{}                                  & 100.0                                                                                       & N/A                                                                                                 \\
                & \romd{}                               & 97.3                                                                                        & 6.3                                                                                                 \\
\textbf{\es{}}     & \es{}                                  & 62.8                                                                                        & 5.9                                                                                                 \\
                & \escr{}                               & 63.0                                                                                        & 6.0                                                                                                 \\
                & \esmx{}                               & 61.5                                                                                        & 5.9                                                                                                 \\
                & \espe{}                               & 60.7                                                                                        & 6.0                                                                                                 \\
                & \esuy{}                               & 61.2                                                                                        & 5.8                                                                                                 \\ \bottomrule
\end{tabular}
\caption{\centering Lexical overlap measures based on the appropriate test data.}
\label{lex}
\end{table}

We introduce a \emph{lexical overlap} metric in order to aid our analysis when comparing results for source-target pairs.  We measure lexical overlap in terms of the overlap of the distinct tokens in the fine-tuning data of the source language and the test data of the target language.  To do so, we apply the subword tokenizer of the source language BERT to the source fine-tuning data and the target test data to obtain the source and target vocabulary sets, and take the intersection.  Given $S$, the vocabulary of the source \emph{fine-tuning} data, and $T$, the vocabulary of the target \emph{test} data, we define lexical overlap as $|S \cap T| \mathrel/ |T|$. The lexical overlap measures are found in Table~\ref{lex}.  Romanian and Moldavian have the highest lexical overlap in the topic identification data, while the Spanish varieties have the lowest lexical overlap in the sentiment analysis data.  Moreover, while the overlap between the fine-tuning and test data in the source language is high for German and a complete match for Romanian, it is low for Spain Spanish, indicating that the sentiment analysis task poses an additional challenge of high lexical variation within the corpus.

To strengthen our comparison of the source and target language, we also introduce a measurement to understand what kinds of tokens are present in the target vocabulary but absent from the source vocabulary. We calculate the average length (in characters) of the out of vocabulary target language tokens. A shorter average length indicates that the tokens from the target data that are not present in the source data are short subwords, while a longer average length indicates that the target data includes longer, more meaningful subwords that are not in the source data.  An example from the data would be the use of ``Alarm'' in German text and ``Wecker'' in Swiss German text; both are in the vocabulary of German BERT; however, because the former is seen more often in association with alarm-related intent labels during fine-tuning, it can be difficult for the model to recognize ``Wecker'' in this context during inference.   
However, character-level noise clearly does not address the ``Alarm'' vs. ``Wecker'' case as there is no surface-level resemblance, so we would not expect to see improvement for language pairs with a longer average length of out-of-vocabulary tokens. We do expect to see improvement for language pairs with a shorter average length of out-of-vocabulary tokens.  

\begin{table*}\centering\small
\begin{tabular}{@{}rlrrrrrrrr@{}}
\toprule
\multicolumn{1}{c}{\textbf{\begin{tabular}[c]{@{}c@{}}Noise\\ Level\end{tabular}}} & \multicolumn{1}{c}{\textbf{\begin{tabular}[c]{@{}c@{}}Comp-\\ osition\end{tabular}}} & \multicolumn{1}{c}{\textbf{\de{}}} & \multicolumn{1}{c}{\textbf{\dech{}}} & \multicolumn{1}{c}{\textbf{\deit{}}} & \multicolumn{1}{c}{\textbf{\nl{}}} & \multicolumn{1}{c}{\textbf{\en{}}} & \multicolumn{1}{c}{\textbf{\da{}}} & \multicolumn{1}{c}{\textbf{\itit{}}} & \multicolumn{1}{c}{\textbf{Average}} \\ \midrule
\textbf{0\%}                                                                       & N/A                                                                                  & 97.6$\pm$0.5                        & 70.7$\pm$3.0                         & 91.5$\pm$2.3                           & 90.9$\pm$1.6                        & \textbf{91.3}$\pm$3.1                        & 82.0$\pm$3.6                        & 73.3$\pm$3.9                        & 85.3$\pm$2.0                             \\
\textbf{25\%}                                                                      & Joint                                                                                & 97.1$\pm$0.8                        & 73.7$\pm$4.0                         & 91.7$\pm$2.5                           & 88.2$\pm$2.4                        & 88.4$\pm$0.4                        & 82.3$\pm$2.7                        & 71.8$\pm$4.1                        & 84.7$\pm$1.4                             \\
\textbf{50\%}                                                                      & Joint                                                                                & 97.8$\pm$0.6                        & 82.1$\pm$2.1                         & 94.0$\pm$2.0                           & 91.2$\pm$1.0                        & 86.1$\pm$4.1                        & 80.3$\pm$2.0                        & \textbf{76.4}$\pm$1.9                        & 86.9$\pm$1.2                             \\
\textbf{75\%}                                                                      & Joint                                                                                & 98.7$\pm$0.4                        & 83.2$\pm$1.7                         & 95.4$\pm$1.3                           & \textbf{92.3}$\pm$1.3                        & 89.4$\pm$2.4                        & 82.3$\pm$3.0                        & 73.4$\pm$3.2                        & 87.8$\pm$0.8                             \\
\textbf{100\%}                                                                     & Joint                                                                                & 98.5$\pm$0.6                        & 83.5$\pm$5.4                         & 96.3$\pm$2.7                           & 89.4$\pm$2.4                        & 89.5$\pm$4.0                        & 82.4$\pm$1.3                        & 74.7$\pm$3.4                        & 87.7$\pm$1.2                             \\
\textbf{25\%}                                                                      & Stacked                                                                              & 98.1$\pm$0.4                        & 80.9$\pm$2.6                         & 95.9$\pm$1.1                           & 91.1$\pm$1.1                        & 90.1$\pm$5.6                        & 85.5$\pm$4.1                        & 69.9$\pm$3.1                        & 87.4$\pm$2.3                             \\
\textbf{50\%}                                                                      & Stacked                                                                              & 98.5$\pm$0.9                        & 86.9$\pm$1.6                         & 96.1$\pm$1.4                           & 88.4$\pm$3.0                        & 87.3$\pm$1.9                        & \textbf{87.1}$\pm$0.8                        & 72.1$\pm$1.5                        & \textbf{88.0}$\pm$0.4                             \\
\textbf{75\%}                                                                      & Stacked                                                                              & \textbf{98.9}$\pm$0.3                        & \textbf{87.3}$\pm$1.8                         & \textbf{96.4}$\pm$1.1                           & 87.9$\pm$3.4                        & 86.5$\pm$2.7                        & 83.5$\pm$2.9                        & 70.3$\pm$1.9                        & 87.2$\pm$1.3                             \\
\textbf{100\%}                                                                     & Stacked                                                                              & 98.8$\pm$0.6                        & 85.7$\pm$3.0                         & 95.5$\pm$1.3                           & 86.6$\pm$3.7                        & 86.1$\pm$3.0                        & 82.9$\pm$3.0                        & 62.5$\pm$7.8                        & 85.4$\pm$2.0                             \\ \bottomrule
\end{tabular}
\caption{\label{intent-mbert}\centering German intent classification results for mBERT with 95\% confidence interval measured for five trials. Bold numbers indicate the highest results (by absolute comparison).}
\end{table*}

\subsection{Nature of the Task}\label{sec:natureofthetask}
The nature of the task seems to dictate the extent to which boosting unseen language performance via noise in fine-tuning is possible.  As described above, success in the intent classification task often comes down to lexical pattern recognition.  For example, sentences in the data might explicitly include ``set alarm to$\ldots$'' when the intent label is set-alarm.  As a result, we are able to reach near-perfect accuracy in the baseline for German (98\%).  However, when it comes to related varieties like Swiss German and South Tyrolean, despite the variations often being small in key intent-related words, the baseline is not able to perform well as it is not robust to such variation.  By including noise in the data, as the results show, we are able to make the model more robust to such variation and see large boosts in performance for all languages. An illustrative example from xSID \cite{van-der-goot-etal-2021-masked} is as follows:

English: Is it going to be sunny today?

German: Wird es heute sonnig?

Swiss German: Isches hüt sunnig?

\noindent The word ``sunny'' is likely enough to cue the model to weather-related intent labels.  In German, it is ``sonnig,'' while in Swiss German, it is ``sunnig.''  This small one-character replacement is enough to change German BERT's subword tokens from ``sonn'' and ``ig'' to ``sun'' and ``nig,'' and because embeddings are tied to tokens, this small difference in spelling can propagate and lead to downstream errors.  Including random character-level noise in fine-tuning helps the model deal with small variations like this.

In contrast, the topic identification and sentiment analysis tasks are difficult to solve simply by surface-level cues.  The baseline performances are indicative of this difficulty: Romanian baseline performance is 77.7\%, and \es{} baseline performance is 66.9\% (as opposed to the near-perfect German and Italian intent classification baseline scores).  Recall that the authors of the MOROCO dataset \cite{butnaru2019moroco} replace all named entities with \$NE\$ placeholders, so it is intentionally made difficult to use surface-level cues for topic identification. Moreover, the low lexical overlap between the fine-tuning and test data for the source Spanish variety (\es{}) is indicative of higher lexical variation within the data, meaning surface-level patterns learned during fine-tuning would not be as helpful at inference.  Though noise makes the model more robust to seeing variations at the \emph{surface} level, these two task settings require deeper cues, so other techniques may be required to further facilitate cross-lingual transfer in such cases.

\subsection{Source-Target Pairs}
The utility of character-level noise for German and Italian intent classification but not Romanian-Moldavian topic identification or Spanish sentiment analysis can be explained in part by the nature of the tasks themselves.  However, we can learn even more by examining the differences in the source-target language pairs.  Examining the lexical overlap measures for the language pairs (Table~\ref{lex}), we see that the pairs with the highest lexical overlap  are Romanian-Moldavian and Italian-Neapolitan, followed closely by the other German- and Italian-sourced pairs.  The Spanish pairs have the lowest lexical overlap.  Lexical overlap leaves open the question of what does not overlap -- we measure this in terms of the average length of the target language tokens that are out of the vocabulary of the source language, as described in Section~\ref{sec:lexicaloverlap}.  Romanian- and Spanish-sourced pairs have higher average lengths, while German- and Italian-sourced pairs have lower average lengths.

Lower lexical overlap paired with high average length suggests that not only does the test data differ substantially from the fine-tuning data, but the differences are in the form of longer subword tokens that could contribute greatly to the meaning of the sentence as a whole.  As described in Section~\ref{sec:lexicaloverlap}, character-level noise can only do so much to help when the differences are on the order of long subword tokens.  As a result, a case where there is lower lexical overlap as well as high average length of out-of-vocabulary (OOV) tokens would not be a good candidate for character-level noise to be used to promote cross-lingual transfer; the example in our experiments is the Spanish sentiment analysis task. 

In contrast, the Romanian-Moldavian pair has an extremely high lexical overlap of 97.3\%, meaning that only 2.7\% of the tokens in the Moldavian test data are out of the vocabulary of the fine-tuning data.  As a result, though this pair also has the highest average length of OOV tokens, it does not pose the same issue as for Spanish because of the low presence of OOV tokens. 

The German- and Italian-sourced pairs strike a happy balance in terms of having a mid- to high-range lexical overlap comparatively, while having the lowest OOV token lengths.  Thus, in addition to the nature of the intent classification task itself being compatible with the character-level noising technique, these specific language pairs possess the ideal properties to see improvement by applying character-level noise. 

\subsection{Monolingual vs. Multilingual}
We focus on the monolingual models for our analysis, as those are the cases in which we truly have zero-shot cross-lingual transfer (target language is not included in the pre-training data for monolingual models).  However, we acknowledge that mBERT can be an effective tool to promote cross-lingual transfer and test our methods on the German intent classification task (one of our success cases) with mBERT for comparison.  We find that multilingual BERT (Table~\ref{intent-mbert}) has a higher baseline score than monolingual German BERT (Table~\ref{intent-de}) for all languages except Swiss German. German, Dutch, English, Danish, and Italian are all included in mBERT's pre-training, contributing to their higher baseline performance.  However, the monolingual German model has a higher baseline score for Swiss German than mBERT. 

For intent classification, our noise intervention boosts the mBERT baseline scores for all language pairs (except English, once again).  The trend of the more noise the better applies here as well; the mBERT model fine-tuned with 100\% noise under the joint composition performs the best across the languages.  Though mBERT achieves better performance on seen languages than German BERT, the Swiss German results demonstrate that German BERT may be better for related but \emph{unseen} varieties.

\section{Conclusion}
In this work, we explore two questions: first, when is it a good idea to use character-level noise in fine-tuning as an agent for zero-shot cross-lingual transfer, and second, in cases where inducing character-level noise is helpful, which noising techniques work the best?  We fine-tune monolingual BERT models on three sentence-level classification tasks, each with a different source language, introducing several variations in the method of noising for the fine-tuning data.  We test on a medley of unseen dialects, closely-related languages, and distant relatives.  We find that one of our test settings lends itself particularly well to our method, while the other two do not.  This distinction comes down to the nature of the task and the relationship (in terms of lexical overlap) between the cross-lingual source-target pair tested.  Our extensions in the space of noising variations allow us to optimize zero-shot cross-lingual transfer to the unseen target languages for the the success case, yielding a boost in performance not only for closely-related pairs, but also for more distant pairs.

\section*{Limitations}
Though we make an effort to maintain the rigor of our methods and analysis, there are some limitations in our approach which could be addressed in future work. First, beyond the nature of the task data itself, a possible reason that character-level noise would not be appropriate for the Spanish sentiment analysis task is that the TASS 2020 dataset contains considerably fewer training examples than the other two tasks’ datasets, so we may not be able to achieve the optimal performance on this task under the BERT fine-tuning paradigm.  In addition, to stay authentic to the raw data, we do not apply any special preprocessing (like removing mentions or hashtags from the Spanish Twitter data); however, it is possible that such factors contribute to success in the task. Furthermore, our analysis involves three dimensions of comparison: the nature of the task, lexical overlap, and average length of out of vocabulary words. To validate our analysis, we would have liked to expand the experiments to incorporate all possible combinations of the three factors; however, we were unable to due to limited availability of task-labeled dialect data. Similarly, though we test several variations of the noising scheme, there are many more possible and we can’t say definitively whether some other character-level noising scheme would work well for the topic identification and sentiment analysis tasks. Finally, we are able to offer anecdotal insight into why introduction of noise contributes to improvements; however, without a formal error analysis we cannot say for sure.  We would like to conduct a thorough error analysis in future. 

\section*{Ethics Statement}
Because our project deals with existing datasets and models, and our method involves synthetic generation of noise, our research process itself does not inherently involve ethical concerns. However, as with any new development, there can always be potential implications of the work that raise ethical concerns. For instance, we discuss methods of applying synthetic noise to text, which could also be used in adversarial attacks. Our method is intended for a zero-shot setting in which a user is using a nonstandard variety related to some standard language. This can be a valuable tool; however, one can imagine a scenario in which a code language is developed un-monitored online communication, but with extensions of our method, performance for a variety of tasks could improve on the code language, enabling undesired monitoring. 

\section*{Acknowledgements}
This material is based upon work supported by the US National Science Foundation under Grant No. IIS-2125948. 

\bibliography{eacl2023}
\bibliographystyle{eacl2023}

\end{document}